\documentclass{bmvc2k}
\usepackage{times}
\usepackage{epsfig}
\usepackage{graphicx}
\usepackage{amsmath}
\usepackage{amssymb}
\usepackage[shortcuts]{extdash}

\newcommand\blfootnote[1]{%
  \begingroup
  \renewcommand\thefootnote{}\footnote{#1}%
  \addtocounter{footnote}{-1}%
  \endgroup
}
\usepackage{url}            
\usepackage{booktabs}       
\usepackage{nicefrac} 

\title{Rethinking Clustering for Robustness}

\addauthor{Motasem Alfarra$^\ast$}{motasem.alfarra@kaust.edu.sa}{1}
\addauthor{Juan C. P\'erez$^\ast\:^{\text{1,}}\:$}{juan.perezsantamaria@kaust.edu.sa}{2}
\addauthor{Adel Bibi}{adel.bibi@eng.ox.ac.uk}{3}
\addauthor{Ali Thabet}{thabetak@fb.com}{4}
\addauthor{Pablo Arbel\'aez}{pa.arbelaez@uniandes.edu.co}{2}
\addauthor{Bernard Ghanem}{bernard.ghanem@kaust.edu.sa}{1}

\addinstitution{
 King Abdullah University of Science and Technology (KAUST)\\
 Saudi Arabia
}
\addinstitution{
 Universidad de los Andes\\
 Colombia\\
}
\addinstitution{
 University of Oxford\\
 United Kingdom\\
}
\addinstitution{
 Facebook Reality Labs (FRL)\\
 Zurich, Switzerland\\
}
\runninghead{M.~Alfarra and J.C~P\'erez \etal}{Rethinking Clustering for Robustness}

\def\ie{\emph{i.e}\bmvaOneDot}
\def\eg{\emph{e.g}\bmvaOneDot}

\def\etal{\emph{et al}\bmvaOneDot}

\newtheorem{proposition}{Proposition}
\begin{document}

\maketitle
\blfootnote{$^\ast$ Equal Contribution.}
\begin{abstract}
This paper studies how encouraging semantically-aligned features during deep neural network training can increase network robustness. Recent works observed that  Adversarial Training leads to robust models, whose learnt features appear to correlate with human perception. Inspired by this connection from robustness to semantics, we study the complementary connection: from semantics to robustness. To do so, we provide a robustness certificate for distance-based classification models (clustering-based classifiers). Moreover, we show that this certificate is tight, and we leverage it to propose \emph{ClusTR} (Clustering Training for Robustness), a clustering-based and adversary-free training framework to learn robust models. Interestingly, \textit{ClusTR} outperforms adversarially-trained networks by up to $4\%$ under strong PGD attacks. Our code for reproducing our results can be found at  \href{https://github.com/clustr-official-account/ClusTR-Clustering-Training-For-Robustness}{https://github.com/rethinking-clustering-for-robustness}.
\end{abstract}

\section{Introduction}
Deep neural networks (DNNs) have demonstrated tremendous success in various fields, from computer vision \cite{imagenet_classification_krizhevsky,long2015fully} and reinforcement learning \cite{human_level_drl,atari_drl} to natural language processing \cite{neural_machine_translation,transformers} and speech recognition \cite{hinton2012deep}. Despite this breakthrough in performance, robustness is becoming a rising concern in DNNs. Specifically, DNNs have been shown to be vulnerable to imperceptible input perturbations~\cite{intriguing,explaining_harnessing}, known as adversarial attacks, which can entirely alter the DNN's output. This vulnerability has popularized a new line of research known as network robustness. Robust DNNs should not only be accurate, but also resistant against input perturbations. Given the importance of the problem, a plethora of network robustness approaches have been proposed, including those based on regularization~\cite{cisse2017parseval,trades,loss_curv_reg_robustness,logit_pairing}, distillation \cite{distillation_papernot}, and feature denoising \cite{feature_denoising}, among many others. In this paper, we focus our attention on the popular and effective adversarial training approach \cite{madryadv}.

Adversarial training explicitly trains DNNs on adversarial attacks generated on-the-fly through projected gradient descent (PGD). This technique has proven to significantly improve network robustness, and has become a standard for training robust networks. Interestingly, and as a byproduct, adversarially-trained networks seem to learn features that are more semantically aligned with human perception~\cite{learning_perceptually_aligned,engstrom2019adversarial}, to such a degree that the learnt DNNs can be used for several image-synthesis tasks \cite{computer_vision_with_single_robust_classifier}.
Learning more semantically-aligned features in DNNs remains an open problem. A promising direction for obtaining features with such properties is through Deep Metric Learning (DML) techniques. DML learns feature representations by preserving a notion of similarity between inputs and their feature representations \cite{deep_met_triplet_net,norouzi_conse}, and has achieved remarkable performance in face recognition \cite{facenet}, image retrieval \cite{frome_dist_fun}, and zero-shot learning \cite{frome2013devise}. The preservation of similarity that DML seeks often involves clustering semantically-similar instances. Hence, recent clustering-based losses \cite{deep_met_triplet_net,magnet} have been designed with this objective in mind, showing significant progress in learning semantic representations that are also competitive in performance with modern classification approaches. 

\begin{figure*}
    \centering
    \includegraphics[width = \textwidth]{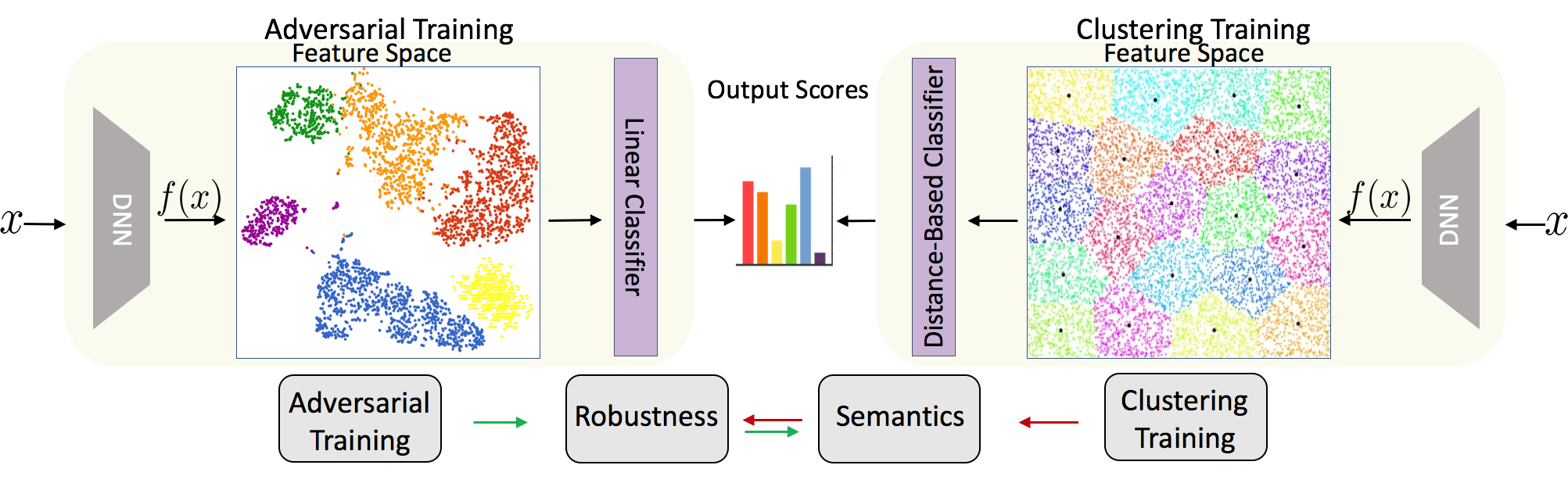}
    \caption{\textbf{Closing the loop on robustness and semantics}. Earlier work showed that adversarial training results in more semantically-aligned features, \ie features of same-class instances tend to cluster together . We study the complementary path, \ie the effect of learning more semantically-aligned features (via clustering) on network robustness (right figure).}
    \label{fig:pipeline}
\end{figure*}

Inspired by these developments, we theoretically show an intimate relation  between semantics (through clustering approaches) and robustness, as illustrated in Figure \ref{fig:pipeline}. In particular, we show that, under certain continuity properties of the DNN, clustering-based classifiers enjoy a tight robustness radius against $\ell_2$-bounded input perturbations. Furthermore, we observe that this radius can be maximized by optimizing a \textit{Clustering Loss}, \ie a loss that encourages clustering of semantically-similar instances in feature space. Inspired by this observation, we show that training DNNs with such a loss results in high-performing classifiers that are also robust against PGD attacks. We enhance this clustering-based approach with standard techniques for DNN training, and dub this framework \textit{Clustering Training for Robustness} (\textit{ClusTR}). To validate the idea behind ClusTR, we experiment on several datasets and find that ClusTR can yield significant robustness gains. In summary, our \textbf{contributions} are three-fold: 
 \textbf{(i)} We study the connection from semantics to PGD robustness by analyzing classifiers that employ clustering in representation space. We use this analysis to derive a tight $\ell_2$ robustness radius, under which all $\ell_2$ perturbations are unable to change the predictions. Moreover, we show that a deep metric learning approach for semantic clustering that optimizes a \textit{Clustering Loss} is directly related to maximizing the derived robustness radius. 
\textbf{(ii)} Motivated by our theoretical findings, we propose the ClusTR framework, which employs a popular \textit{Clustering Loss} (the \textit{Magnet Loss} \cite{magnet}), to learn robust models against PGD attacks without generating adversaries during training.
We validate the theory behind ClusTR through extensive experiments and find that ClusTR results in a significant boost in robustness against PGD attacks without relying on adversarial training. Specifically, we observe that classifiers learnt using ClusTR outperform (in robustness) adversarially-trained classifiers \cite{freeadv}
by $3\%$ and $4\%$ under strong $\nicefrac{8}{255}$ PGD attacks on the CIFAR10 \cite{cifars} and SVHN~\cite{svhn} datasets, respectively.\textbf{(iii)} Equipping ClusTR with a quick and cheap version of adversarial training can increase robustness against $\nicefrac{8}{255}$ attacks on several benchmarks by significant margin.

\section{Related Work}
\textbf{Metric Learning.} The idea of encouraging learnt features to be more semantically meaningful to the human visual system has been extensively studied in the metric learning community, where the goal is to learn a similarity measure in feature space that correlates with a similarity measure between inputs~\cite{dist_metric_learning_clustering,dml_neighbor,deep_met_triplet_net,dml_survey,zhu2019new}. In such a setting, semantically-similar inputs (\ie those belonging to the same class) are expected to be clustered together. This paradigm has shown remarkable performance in several tasks \cite{facenet,mikolov2013distributed,frome_dist_fun}. Closely related to our work, the approach of~\cite{mao2019metric} used the \textit{Triplet Loss}~\cite{facenet} to regularize learnt features and enhance network robustness. We complement the previous art with a theoretical justification on the intimate relation between robustness and the general family of metric-learning classifiers that subsumes the \textit{Triplet Loss} as a special case. Namely, we find a connection between the \textit{Magnet Loss} \cite{magnet} and theoretical guarantees of network robustness.

\noindent \textbf{Adversarial Robustness.}
The existence of adversarial perturbations has dramatically increased security concerns in DNNs. Consequently, there has been a surge of research aiming at learning adversarially-robust models \cite{buckman2018thermometer,ma2018characterizing,cisse2017parseval}. Despite its high computational cost, adversarial training \cite{madryadv} remains one of the most popular, successful and reliable techniques for attaining adversarial robustness. Furthermore, adversarial training was regularized by enforcing similarity between logits of both natural and adversarial pairs \cite{logit_pairing}. This work was further developed in TRADES~\cite{trades}. Moreover, regularization also studied the data-complexity perspective, demonstrating an inherent sample complexity barrier on robust learning \cite{more_data_robustness}, and that pre-training or learning from unlabeled data can vastly improve robustness of adversarially-trained networks \cite{pretraining, carmon2019unlabeled}.

\noindent \textbf{Robust Features.} Recent work demonstrated that networks trained adversarially enjoy an unexpected benefit: the learnt features tend to align with salient data characteristics and human perception~\cite{robustness_odds_acc}. Moreover, the learnt features, commonly referred to as robust features \cite{adversarial_ara_not_bugs}, seem to be clustered in feature space, while being perceptually aligned \cite{learning_perceptually_aligned}. Based on these findings, the power of such semantically-aligned features was harnessed to perform image synthesis tasks with a single robust classifier \cite{computer_vision_with_single_robust_classifier}. In this paper, we take an orthogonal direction to robustness, in which we encourage robustness by training DNNs to specifically learn more semantically-aligned features via clustering.

\begin{figure*}[t]
    \centering
    \includegraphics[width=0.9\textwidth]{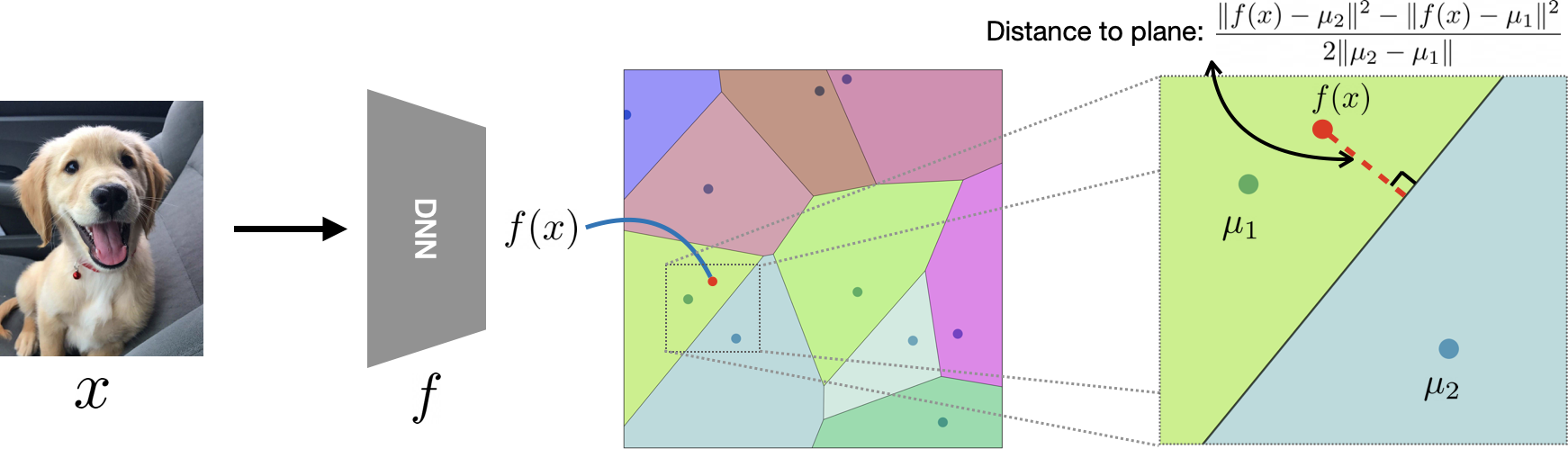}
    \caption{\textbf{Illustration of Proposition \ref{thm1}}. For a classifier $f$ trained with a \textit{Clustering Loss}, an instance $x$ is classified by assigning it to the class of the closest cluster to its feature representation $f(x)$. The resulting decision boundaries form a Voronoi diagram in feature space. As a consequence, the robustness radius in Proposition \ref{theo:delta_bound} is proportional to the distance to the decision boundary separating the two closest clusters to $f(x)$.}
    \label{fig:theorem1_pull_figure}
\end{figure*}

\section{From Robustness to Clustering Loss}
Recent work has shown that adversarially-trained DNNs, while robust, also tend to learn more semantically-aligned features \cite{robustness_odds_acc, learning_perceptually_aligned}. Inspired by these findings, we are interested in studying the converse implication, \ie whether DNNs trained to learn such features enjoy robustness properties. To this end, we start by studying the robustness of a common family of classifiers used in deep metric learning~\cite{deep_met_triplet_net,magnet},  namely classifiers that are based on clustering semantically-similar inputs.

\subsection{Robustness}
\textbf{Clustering-based classifiers.} Consider a training set consisting of input-label pairs $\mathcal{D} = \{x_i,y_i\}_{i=1}^N$, where $x_i \in \mathbb{R}^n$ belongs to one of $L$ classes, and a parameterized function $f_\theta : \mathbb{R}^n \rightarrow \mathbb{R}^d$, which can be a DNN. A clustering-based classifier learns parameters $\theta$ such that $f_\theta$ clusters semantically-similar inputs $x_i$ (inputs with similar labels $y_i$) in feature space $\mathbb{R}^d$. That is, $f_\theta$ clusters each of the $L$ classes into $K$ different clusters (where $K$ may vary across classes). Hence, an input $x_i$ is assigned a label $c$, if and only if, $f_\theta(x_i)$ is closest, under some notion of distance, to one of the $K$ clusters representing class $c$. To analyze the robustness of such classifiers, and without loss of generality, we consider a binary classification problem, where inputs belong to one of two classes, $\mathcal{C}_1$ or $\mathcal{C}_2$, and each class is represented with a single cluster center, \ie $L=2$ and $K=1$. Let the cluster centers of $\mathcal{C}_1$ and $\mathcal{C}_2$ be $\mu_1$ and $\mu_2$, respectively, in $\mathbb{R}^d$. Thus, $x_i$ is classified as $\mathcal{C}_1$, if and only if, $\|f_\theta(x_i) - \mu_1\| < \|f_\theta(x_i) - \mu_2\|$, and as $\mathcal{C}_2$ otherwise. Throughout this paper, we assume that $f_\theta$ is $\mathcal{L}_f$-Lipschitz continuous~\cite{cisse2017parseval}, \ie $\|f_\theta(x) - f_\theta(y)\| \leq \mathcal{L}_f\, \|x-y\| ~\forall x,\:y$, where $\|\cdot\|$ denotes the $\ell_2$ norm.

We are interested in the maximum norm of an input perturbation $\delta$ such that the clustering-based binary classifier assigns the same class to both $x$ and $(x+\delta)$. The following proposition provides a bound on such a $\delta$, denoted as the robustness radius.

\begin{proposition} \label{thm1}
Consider the clustering-based binary classifier that classifies $x$ as class $\mathcal{C}_1$, \ie $\|f_\theta(x) - \mu_1\| < \|f_\theta(x) - \mu_2\|$, with $\mathcal{L}_f$-Lipschitz $f_\theta$. The classifier's output for the perturbed input $(x+\delta)$ will not differ from $x$, \textit{i.e.} $\|f_\theta(x+\delta) - \mu_1\| < \|f_\theta(x+\delta) - \mu_2\|$, for all perturbations $\delta$ that satisfy:
\begin{align}
\label{theo:delta_bound}
    \|\delta\| <\,\, \frac{\|f_\theta(x) - \mu_2\|^2 - \|f_\theta(x) - \mu_1 \|^2}{2 \mathcal L_f \|\mu_2 - \mu_1\|}.
\end{align}
\end{proposition}
\emph{Proof Sketch.}
It suffices to observe that the clustering-based classifier is equivalent to a linear classifier, operating in representation space, defined by the hyperplane $(\mu_1 - \mu_2)^\top (f_\theta(x) - \nicefrac{(\mu_2 + \mu_1)}{2}) = 0$. The result is deduced from the Cauchy-Schwarz inequality and the Lipschitz continuity property of $f_\theta$, where the bound is proportional to the $\ell_2$ distance to the hyperplane, as illustrated in Figure \ref{fig:theorem1_pull_figure}.

\textbf{Generalization to the Multi-Class Multi-Cluster Setting.} We first consider the multi-class single-cluster case, \ie $L \ge 2$, $K = 1$, where each class is represented by a single cluster center $\mu_i$, as depicted in Figure \ref{fig:theorem1_pull_figure}. Analyzing the robustness around an input $x$ in this case is equivalent to analyzing the previously discussed binary classification case with respect to the two closest cluster centers \ie $\mu_1 = \mu_{i^*} = \text{arg}\min_{i \in \{1,\dots,L\}} \|f_\theta(x) - \mu_i\|$ and $\mu_2 = \mu_{j^*} = \text{arg}\min_{i \in \{1,\dots,L\}/\{i^*\}} \|f_\theta(x) - \mu_i\|$. We leave the rest of the details for the \textbf{appendix}.

\subsection{Clustering Loss as a Robustness Regularizer}
Proposition \ref{theo:delta_bound} provides a tight robustness radius for each input. To attain both accurate and robust models, one can train DNNs to achieve accuracy, while simultaneously maximizing the robustness radius in Proposition \ref{thm1} for every training input $x$. Several observations can be made about the robustness radius. First, it is inversely proportional to the DNN's Lipschitz constant $\mathcal{L}_f$, \ie networks with smaller $\mathcal{L}_f$ tend to enjoy better robustness. This is consistent with previous work that exploited this observation to enhance network robustness~\cite{cisse2017parseval}. In this paper, we focus on the term $\|f_\theta(x) - \mu_2\|^2 - \|f_\theta(x) - \mu_1\|^2$, and on learning parameters $\theta$ to maximize it, \ie to push features far from cluster centers of different classes ($\mu_2$) and to pull features closer to cluster centers of their class ($\mu_1$). As such, a general class of robustness-based clustering losses can be formulated as follows:
\begin{equation}\label{eq:second_loss}
    \begin{aligned}
    \mathcal{L}^{\text{Robust}}_{\text{Clustering}} = \frac{1}{N}\sum_{i=1}^N \mathcal H\Bigg( & \mathcal F\Big( f_\theta(x_i),  \{\mu_{c_i,j}\}_{j=1}^K\Big), \mathcal G \Big(  f_\theta(x_i) , \{\mu_{v\neq c_i,j}\}_{j=1}^K \Big) \Bigg),
    \end{aligned}
\end{equation}

\noindent where $c_i = \mathcal{C}(x_i)$ is the class of $x_i$ and $\mu_{i,j}$ denotes the $j^{\text{th}}$ cluster of class $i$. The function $\mathcal{F}$ measures the separation between the feature representation of $x_i$, \ie $f_\theta(x_i)$, and the cluster centers of its class. Similarly, $\mathcal{G}$ measures the separation between $f_\theta(x_i)$ and the cluster centers of all other classes. The function $\mathcal H$ combines the two measurements in an overall stable loss, so that minimization of the loss incites larger values for the numerator in Proposition \ref{theo:delta_bound}. 
Note that iterative optimization of this loss requires updating $\theta$. Hence, after every update, cluster centers $\mu_{i,j}$ can be recomputed by any clustering algorithm, \eg \emph{K-means}. Moreover, many losses commonly used in the deep metric learning literature \cite{NCM} conform with Equation \eqref{eq:second_loss} as special cases, one of which is the popular \textit{Magnet Loss}~\cite{magnet}, defined as:
\begin{equation}\label{eq:magnet_loss}
\begin{aligned}
    \mathcal{L}_\text{Clustering}^\text{Magnet} = \frac{1}{N}\sum_{i=1}^N &  \Bigg\{ \alpha + \frac{1}{2\sigma^2}\|f_\theta(x_i) - \mu_{c_i,v^*}\|^2 +  \log\left(\sum_{j=1}^K \sum_{v \neq c} e^{-\frac{1}{2\sigma^2}\|f_\theta(x_i) - \mu_{v,j}\|^2}\right)\Bigg\}_+
\end{aligned}
\end{equation}

\noindent where $\{x\}_+ = \max(x,0)$, $\sigma^2= \frac{1}{N-1} \sum_{i=1}^N \|f(x_i) - \mu_{c_i,v^*}\|^2$, $\alpha \geq 0$, and $v^* = \text{argmin}_v \|f_\theta(x_i) - \mu_{c_i,v}\|$. Note that the \emph{Magnet Loss} is a special case of the previously formulated general Robust Clustering loss that incentivizes the increase in the numerator of Proposition \ref{thm1}. That is to say, the feature representations $f_\theta(x_i)~\forall i$ are pulled closer to clusters representing the correct class and pushed away from clusters of other classes. While the \emph{Magnet Loss} was introduced to address performance issues in metric learning algorithms, our objective of learning more semantically-aligned features and our subsequent analysis of Proposition \ref{theo:delta_bound} suggest that this loss \textit{inherently} encourages robustness. 

\noindent Regarding inference, DNNs trained with \emph{Magnet Loss} predict the class of a test input by computing a soft probability over the features produced by $f_\theta$, as follows:
\begin{equation}\label{eq:inference_ours}
\begin{aligned}
Pr(\mathcal C(x_i) = c ) = p_c(f_\theta(x_i)) = \frac{\sum_{j=1}^K  e^{-\frac{1}{2\sigma^2} \|f_\theta(x_i) - \mu_{c,j}\|^2}}{\sum_{j=1}^{K} \sum_{v=1}^L e^{-\frac{1}{2\sigma^2} \|f_\theta(x_i) - \mu_{v,j}\|^2}}.
\end{aligned}
\end{equation}
Hence, $x_i$ is assigned to class $\text{argmax}_{c}\, p_c(f_\theta(x_i))$. We refer the  reader to~\cite{magnet} for more details. 

\subsection{ClusTR: Clustering Training for Robustness}\label{sec:clustr}
Our theoretical study finds an intrinsic connection between clustering and robustness: clustering\=/based classifiers intrinsically possess a robustness radius. As such, optimizing a loss designed for clustering tends to maximize this robustness radius. We also observe that a \emph{Clustering Loss} such as Equation \eqref{eq:second_loss}, which is designed to induce robustness according to Proposition \ref{thm1}, can be reduced to the \textit{Magnet Loss} of Equation \eqref{eq:magnet_loss} as a special case. Based on these observations, we propose Clustering Training for Robustness (ClusTR): a simple and theoretically-motivated framework for inducing robustness during DNN training without the need to generate adversaries. ClusTR exploits our theoretical findings by combining a \emph{Clustering Loss} with simple DNN-training techniques. 

For the \emph{Clustering Loss}, ClusTR incorporates the well-studied \textit{Magnet Loss} to induce semantic clustering of instances in feature space. Although effective in its task, this loss suffers from slow convergence~\cite{magnet}. ClusTR mitigates this issue by introducing a simple \emph{warm start} initialization. For a given model and dataset, ClusTR first conducts nominal training, \ie standard Cross Entropy training, until reasonable performance is achieved. Then, it removes the last linear layer and fine-tunes the resulting DNN by applying the \textit{Magnet Loss} on the output of the penultimate layer. The \textit{Magnet Loss} in ClusTR aims at optimizing the robustness radius of Proposition \ref{thm1}, while using a warm start initialization to increase convergence speed without hindering test set accuracy.  In this work, we choose the \textit{Magnet Loss} to be the \textit{Clustering Loss} in ClusTR. However, we remark that Proposition~\ref{theo:delta_bound} is agnostic to this choice, so we expect our results to extend to other choices of a \textit{Clustering Loss}.

\section{Experiments}\label{sec:experiments}

\begin{figure*}
    \centering
\includegraphics[width = \textwidth]{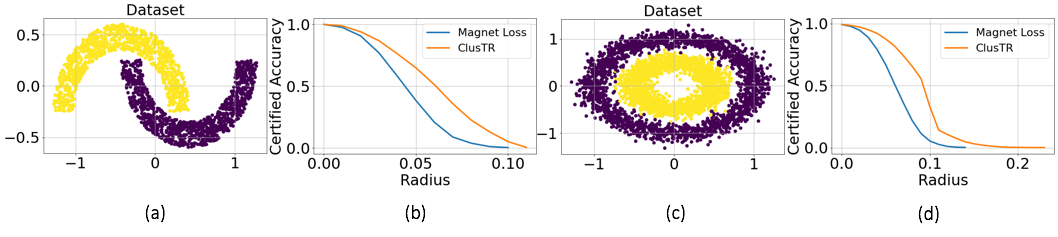}
    \caption{\textbf{Effect of \emph{warm start} on certified accuracy}. Figures (a)-(c) show the synthetic datasets, while Figures (b)-(d) show the effect of \emph{warm start} in ClusTR on certified accuracy. In both datasets, \emph{warm start} induces a larger robustness radius than  random initialization.}
    \label{fig:synthetic}
\end{figure*}

In this section, we conduct several experiments on synthetic and real datasets to validate the idea behind ClusTR. 
Specifically, we study (a) the effect of a \textit{warm start} on convergence speed and robustness, (b) how ClusTR-trained DNNs compare to their adversarially-trained counterparts, and (c) how ClusTR can be equipped with a quick version of adversarial training to further enhance robustness.

\subsection{Effect of Warm Start Initialization in ClusTR}

\textbf{Convergence.} We assess the training convergence and the overall test accuracy performance for our proposed ClusTR-training of ResNet18 on CIFAR10 and SVHN. In CIFAR10, we observe that training without warm start (\ie \textit{Magnet Loss} only) requires 106 minutes to fully train, while introducing the warm start reduces the required training time to 83 minutes.

\noindent \textbf{Robustness.} We study the effect of the warm start initialization on robustness by conducting controlled synthetic experiments and computing \textit{exact} robustness radii by computing a tight estimate of the Lipschitz constant. We train a 3-layered neural network with 20 hidden units on the synthetic binary classification datasets depicted in Figures \ref{fig:synthetic}(a) and (c). On both datasets, we train (1) \textit{Magnet Loss} with random initialization and (2) ClusTR. For simplicity, each class is represented with a single cluster, \ie $K=1$.
Upon convergence, both models achieve $99\%$ accuracy.
Given model predictions, we compute the robustness radius for each instance and report certified accuracy under various radii $r$ in Figures \ref{fig:synthetic}(b) and (d). This is in line with common practice in the network certification literature \cite{randomized_smoothing}. Note that certified accuracy at radius $r$ is defined as the percentage of instances that are both correctly classified and have a robustness radius larger than $r$, as given by Proposition \ref{theo:delta_bound}. 
We find that the ClusTR-trained DNNs, while accurate, also enjoy a larger robustness radius than DNNs trained with \textit{Magnet Loss} without the warm start.

\subsection{ClusTR Robustness against PGD}

\textbf{Setup and Implementation Details.} In this section, we conduct experiments with ResNet18 on the CIFAR10, CIFAR100, and SVHN datasets. We train models using our proposed ClusTR framework. Specifically, we first conduct nominal training until we get a reasonable performance\footnote{Models with test accuracies of $90\%, 75\%, 90\%$ on CIFAR10, CIFAR100 and SVHN, respectively.}. We then remove the last linear layer and fine-tune the network by applying the \textit{Magnet Loss} on the output feature of the resulting DNN. Fine-tuning is done for 30 epochs on CIFAR10 and SVHN, and 60 epochs on CIFAR100. 
Following \cite{magnet}, we use $k$-means$_{++}$~\cite{kmeans++} to update cluster centers after each training epoch. To assess model robustness, we follow prior work and perform projected gradient descent (PGD) \cite{madryadv} attacks with $\epsilon$-$\ell_\infty$-bounded perturbations that take the following form:
\begin{equation}
\begin{aligned}
x^{k+1} = \prod_{\mathcal{S}}\left(x^k + \eta \: \text{sign}\left(\nabla_{x^k}\mathcal L_{\text{ce}}(p(f_\theta(x^k)),y)\right)\right),
\end{aligned}
\end{equation}
where $\prod_{\mathcal{S}}(x+\delta)$ denotes the projection of the perturbed input onto the set $\mathcal{S} = \{(x+\delta)\in[0,1]^n, \|\delta\|_\infty\leq\epsilon\}$, $p(f_\theta(x^k))$ is the probability prediction vector computed through Equation \eqref{eq:inference_ours}, and $\mathcal L_{\text{ce}}$ is the Cross Entropy loss. In all experiments, we perform PGD attacks with 10 random restarts around each input for 20 and 100 iterations, denoted as PGD$^{20}$ and PGD$^{100}$, respectively. Following common practice in the literature \cite{freeadv,fastadv}, we set the PGD step size to $\eta = \nicefrac{2}{255}$. We report the attacks with an attack strength of $\epsilon = \nicefrac{8}{255}$, and leave experiments with other choices of $\epsilon$ for the \textbf{appendix}.

\begin{table*}[t]
\small
\centering
\caption{\textbf{Adversarial accuracy comparison on CIFAR10 and SVHN}. We compare ClusTR and ClusTR+QTRADES against Magnet Loss, Free Adversarial Training (Free AT), AT with ImageNet pre-training, TRADES, and QTRADES under $\epsilon = \nicefrac{8}{255}$ PGD attacks. ClusTR+QTRADES outperforms the adversarially-trained models by a large margin.}

\centering
\begin{tabular}{l | cccc  ccc}
\hline \bottomrule  
 & \multicolumn{3}{c}{CIFAR10} & \multicolumn{3}{c}{SVHN} \\
                                            & Natural       & PGD$^{20}$    & PGD$^{100}$   &  Natural      & PGD$^{20}$    & PGD$^{100}$ \\ \midrule
\text{Nominal Training}                     & \textbf{95.01}& 0.00          & 0.00          & \textbf{98.38}& 0.00          & 0.00 \\\hline
\text{Free AT \cite{freeadv}}               & 85.96         & 46.33         & 46.19         & 86.98         & 46.52         & 46.06 \\
\text{AT + Pre-Training \cite{pretraining}} & 87.30         & 57.40         & 57.20         & 85.12         & 47.18         & 46.72 \\
\text{TRADES \cite{trades}}                 & 84.92         & 56.61         & 56.43         & 91.63         & 57.45         & 55.28 \\\hline
\text{Magnet Loss \cite{magnet}}            & 83.14         & 23.71         & 22.54         & 91.95         & 40.73         & 38.59 \\
\text{ClusTR}                               & 87.34         & 49.04         & 47.76         & 94.28         & 50.78         & 50.77 \\
\text{QTRADES}                              & 81.07         & 44.18         & 43.42         & 86.36         & 43.05         & 42.24 \\
\text{ClusTR + QTRADES}                     & 91.03         & \textbf{74.44}& \textbf{74.04}& 95.06         & \textbf{84.76}& \textbf{84.75} \\
\hline \bottomrule      
\end{tabular}
\label{tb:best_results_svhn_and cifar10}
\end{table*}

\noindent \textbf{Experiments on CIFAR10 and SVHN.} We evaluate the PGD robustness of nominal training (as baseline), the Magnet Loss (\ie ClusTR without warm start), and ClusTR, and we compare against several approaches that provide PGD robustness in this experimental setup, namely \textit{Free adversarial training} (Free AT) \cite{freeadv} with its reported best setting of 8 minibatch-replays that outperforms vanilla adversarial training \cite{madryadv}, \textit{Adversarial Training with ImageNet pre-training} (AT + PreTraining) that leverages external data to improve robustness, and \textit{TRADES} \cite{trades}. Note that all the robustness methods in this comparison employ various forms of adversarial training. 
We report both natural accuracy, \ie test set accuracy on clean images, and PGD test accuracy. Table~\ref{tb:best_results_svhn_and cifar10} reports these results. First, we observe that training with Magnet Loss \textit{only} on clean images results in substantial gains in robustness compared to nominal training. In fact, this choice of loss function increases PGD$^{20}$ accuracy from 0\% to 23.71\%, while natural accuracy drops from 95.01\% to 83.14\%. This result constitutes empirical evidence of the theoretical robustness properties we presented for clustering-based classifiers. Furthermore, training with ClusTR consistently outperforms Free AT in both natural and PGD accuracy for both CIFAR10 and SVHN. Specifically, ClusTR outperforms Free AT in PGD$^{20}$ accuracy by $3\%$ and $4\%$ on CIFAR10 and SVHN, respectively, even though the former \textit{only} trains with clean images. 
We note that ClusTR's robustness gains over adversarial training are not accompanied with lower natural accuracy. In fact, the natural accuracy of ClusTR is 1\% more in CIFAR10 and 7\% more in SVHN.
These results show that the design of ClusTR inherently provides robustness properties without introducing adversaries during training. We complement this finding by studying the following question: Can equipping ClusTR with some form of adversarial training provide even larger PGD robustness gains? We equip ClusTR with a TRADES loss term, where the total loss becomes: 
\begin{equation}\label{eq:total loss}
    \mathcal L_{\text{Total}} = \mathcal{L}^{\text{Magnet}}_{\text{Clustering}} + \lambda \mathcal{L}_{\text{ce}}(p(f_\theta(x_{\text{adv}})),p(f_\theta(x))).
\end{equation}
Note that the Cross Entropy-based TRADES formulation \cite{trades} is similar to Equation \eqref{eq:total loss}, but with the first term replaced with $\mathcal{L}_{\text{ce}}(p(f_\theta(x)),y)$, where $p(f_\theta(x))$ is the output logits of the last linear layer and $y$ is the true label. In order to keep the framework simple and computationally efficient, we compute a \textit{quick} estimate of the adversary $x_{\text{adv}}$ in Equation \eqref{eq:total loss}. Namely, we start from a random uniform initialization and perform a \textit{single} PGD step as opposed to TRADES' multiple iterations. We refer to this setup as QTRADES\footnote{The rest of the implementation details of QTRADES are left for the \textbf{appendix}.}. Formally, for an input $x$, we construct an adversary by perturbing $x$ with uniform noise, \ie $x'= x + \mathcal{U}[-\epsilon, \epsilon]$, and then generate $x_\text{adv}$ by:
\[
x_{\text{adv}} = \prod_{\mathcal{S}}\left(x' + \eta \: \text{sign}\left(\nabla_{x'}\mathcal L_{\text{ce}}(p(f_\theta(x')),p(f_\theta(x)))\right)\right).
\]
We report results for this experiments in Table \ref{tb:best_results_svhn_and cifar10}. While QTRADES alone only achieves slightly lower natural accuracy and adversarial robustness (when compared to Free AT), our results show that equipping ClusTR with QTRADES enhances PGD robustness results on both datasets, outperforming all other methods. In particular, we observe that ClusTR+QTRADES achieves the highest natural accuracy among all methods with $91.03\%$ and $95.06\%$, on CIFAR10 and SVHN, respectively, thus improving upon the best competitor by $4\%$ on both datasets. Also, ClusTR+QTRADES surpasses other baselines by sizable margins: $16.84\%$ and $29.47\%$ under strong PGD attacks on CIFAR10 and SVHN, respectively.

\noindent \textbf{Experiments on CIFAR100.} We extend our analysis of ClusTR+QTRADES to CIFAR100, and assess PGD robustness with $\epsilon = \nicefrac{8}{255}$ attacks. We report the results of this setup in Table~\ref{tb:best_results_cifar100}, which shows that ClusTR+QTRADES outperforms the strongest competitor by $18.25\%$ under strong PGD attacks. We note that these large gains in PGD robustness also come with a substantial $7\%$ increase in natural accuracy. For CIFAR100, the total number of clusters is $100 \text{ (classes)} \times 2 \text { (clusters per class)} = 200$. Following how \cite{magnet} tackles the large-cluster-number regime, in this case we compute predictions for ClusTR+QTRADES  without considering all clusters, as in Equation \eqref{eq:inference_ours}, but only the $D$ nearest clusters. While we take  $D = 20$ in this experiment, we find that the choice of $D$ around this value has a marginal impact on robustness. We leave an ablation of $D$ for the \textbf{appendix}.

\begin{table}[t]
\small
\centering
\caption{\textbf{Adversarial accuracy on CIFAR100}. We compare ClusTR+QTRADES against Free AT, AT+Pre-Training, and TRADES under $\epsilon = \nicefrac{8}{255}$ PGD attacks. Our proposed ClusTR+QTRADES framework surpasses all competition by a large margin.}
\centering
\begin{tabular}{l| ccc}
\hline \bottomrule 
 & \multicolumn{3}{c}{CIFAR100} \\
     & Natural & PGD$^{20}$ & PGD$^{100}$ \\
\midrule
\text{Nominal Training}         & \textbf{78.84} & 0.00 & 0.00 \\\hline
\text{Free AT \cite{freeadv}}     & 62.13 & 25.88 & 25.58  \\
\text{AT+Pre-Training \cite{pretraining}}          & 59.23 & 34.22 & 33.91 \\
\text{TRADES \cite{trades}}                   & 55.36 & 28.11 & 27.96 \\\hline
\text{ClusTR+QTRADES}              & 69.25 & \textbf{52.47} & \textbf{52.40} \\
\hline \bottomrule  
\end{tabular}
\label{tb:best_results_cifar100}
\end{table}

\noindent \textbf{Adaptive Attacks}. While going against the current paradigm in the network robustness literature, it has been argued that common attacks may be insufficient to demonstrate network robustness. Specifically, recent work shows that many defenses can be broken with carefully-crafted attacks \cite{obfus}, now dubbed \textit{adaptive attacks}, tailored to break the underlying defense \cite{adaptive_attacks}. 
Following this principle, we construct a potential powerful attack tailored to our trained networks. Namely, we construct adversaries that maximize the $\mathcal{L}^{\text{Magnet}}_{\text{Clustering}}$ loss, as opposed to the standard Cross Entropy loss in the PGD formulation. Similar to previous experiments, the attacks are performed with 10 random restarts for 100 iterations and $\epsilon = \nicefrac{8}{255}$. Note that this attack precisely targets the objective, with which our models are trained, thus, the attack is expected to be stronger. Indeed, running this adaptive attack lowers the robustness accuracy from $74.04\%$ to $66.52\%$ on CIFAR10, and from $84.75\%$ to $78.79\%$ on SVHN. Despite this drop, our ClusTR+QTRADES approach still outperforms other methods by substantial margins. It is essential to note here that this drop in robustness is considered to be rather marginal, as other defenses, when subjected to such tailored attacks, have their robustness drop close to 0, or at least to lower-than-baseline robust models \cite{adaptive_attacks, obfus}.

\noindent It is worthwhile to mention that our choice of QTRADES, out of the many adversarial training schemes with which ClusTR can be equipped, is motivated by \textit{(i)} the theoretical support behind TRADES \cite{trades} and \textit{(ii)} QTRADES' low computational cost. We also emphasize here that PGD robustness could possibly be improved further by incorporating another adversarial training technique with ClusTR instead of QTRADES. We leave the search for this optimal choice to future work. 

\subsection{Discussion}
Evaluating adversarial robustness is a complex task with frequent methodological changes~\cite{carlini2019evaluating}. Empirically evaluating defense mechanisms requires selecting a threat model and an adversary that aims at exploiting such threat. In this work, we theoretically characterized an existing connection between semantics, achieved through clustering, and adversarial PGD robustness. Given our theoretical insights, we then set out to conduct proof-of-concept experiments to test our findings. For this purpose, we choose the threat model of $\ell_\infty$ attacks of norm smaller than a given $\epsilon$, and the well-studied Cross Entropy-based PGD adversary to conduct attacks. Our experiments show that networks trained with our approach are notably robust against this adversary. These results provide empirical evidence that our theoretical findings correspond with practical applications. However, we abstain from claiming that our approach provides adversarial robustness in the most generic sense beyond PGD attacks: other attacks may be able to find vulnerabilities in our defense.

\noindent \textbf{Acknowledgments.}
This work was  supported by the King Abdullah University of Science and Technology (KAUST) Office of Sponsored Research (OSR) under Award No. OSR-CRG2019-4033.
\bibliography{egbib}
\newpage
\appendix

\section{Implementation Details}
We describe the implementation details of ClusTR, along with details regarding QTRADES.

\noindent \textbf{Architecture.} We use a ResNet18 \cite{resnets} modified to accept $32 \times 32$ input images. 
The size of the output of the network in the penultimate layer, \ie the feature dimension, is set to $512$ for all experiments.

\noindent \textbf{Optimization.} For the warm start stage of training ClusTR, we use the Adam optimizer \cite{kingma2014adam} for 90 epochs with learning rate of $10^{-2}$ that is multiplied by $10^{-1}$ at epochs 30 and 60 with cross entropy loss. After that, we fine-tune the DNN with the Magnet Loss with a learning rate of $10^{-4}$ for another 30 epochs for CIFAR10 and 60 epochs for CIFAR100 and SVHN.

\noindent \textbf{Pre-processing.} Images are normalized by their channel-wise mean and standard deviation. For CIFAR10 and CIFAR100. We apply standard data augmentation of random $32\times32$ crops with a padding of 4. For SVHN, we do not employ any data augmentation.

\noindent \textbf{Magnet Loss.} Following Rippel \textit{et al.} \cite{magnet}, we compute a stochastic approximation of the \textit{Magnet Loss}. Hence, Magnet Loss training requires sampling neighborhoods of points in representation space, rather than independent samples. These neighborhoods are defined by a number of clusters and a number of samples per cluster. This sampling procedure does not guarantee that every instance will be sampled, nor that an instance shall be sampled only once. Therefore, we define an epoch as passing as many instances as there are available in the dataset, regardless if some instances were repeated or some instances were seen more than once. We use $K=2$ as the number of clusters per class for our experiments. For sampling, we set the total number of sampled clusters to 12, and the number of samples per cluster to 20. Hence, the total amount of samples in each batch of each batch is $12 \times 20 = 240$. Cluster assignments are recomputed at the end of every epoch with the \textit{K-means} clustering algorithm with the \textit{K-means}$_{++}$ initialization. We run grid search for optimizing the $\alpha$ parameter in the Magnet Loss. We set $\alpha$ to $12.5$ for ClusTR and ClusTR+QTRADES on CIFAR10; to $13$ for ClusTR and to $10$ for ClusTR+QTRADES on SVHN; to $8.5$ for ClusTR+QTRADES on CIFAR100.

\noindent \textbf{QTRADES.} We initialize the adversary by adding uniform noise in $[-\epsilon, \epsilon]$ to the original instance, computing Cross Entropy between the original and adversarial instances and following one step of gradient ascent for Cross Entropy. The result of gradient ascent is always clipped so that the adversarial instances lies in image space, \ie $[0, 1]^n$. The total loss with which the network is trained is a weighted sum of the Clustering Loss and the Cross Entropy between the original and adversarial instances. We cross validate over the regularization term $\lambda$ balancing the two terms in Equation \eqref{eq:total loss}. We set $\lambda$ to $8$ on CIFAR10, to $9.7$ on SVHN, and to $2$ on CIFAR100.

\section{Additional Experiments}

\subsection{Combining CE with Distance-Based Classifier}
The robustness radius in Proposition \ref{theo:delta_bound} holds for any clustering-based classifier of features produced by a Lipschitz-continuous function $f_\theta$. Therefore, we start by addressing the following question: if robustness is the aim, can one replace the last layer of a nominally-trained DNN with a clustering-based classifier to achieve robustness? Addressing this question is essential to establish the necessity of enforcing clustering during training, \ie training with ClusTR. To answer this question, we study a nominally-trained ResNet18 on CIFAR10, which achieves an accuracy of $95.0\%$. We observe that directly applying \textit{K-means} on the representations of the penultimate layer, and performing classification according to Equation \eqref{eq:inference_ours} achieves an accuracy of $21.6\%$, \ie a performance drop of over $70\%$. As adversaries will aim at changing the classifier's predictions, the highest adversarial accuracy that this classifier can attain is upper bounded by $21.6\%$. 
This result demonstrates that features learnt through nominal training are not spatially configured for clustering-based classification. Hence, this result establishes that exploiting the benefits of clustering-based classification requires to explicitly enforce clustering during DNN training.

\subsection{Results of PGD Attacks with Other $\epsilon$ Values.}

\begin{table*}[t]
\centering
\caption{\textbf{Performance of ClusTR+QTRADES on CIFAR10, CIFAR100 and SVHN}. We report the PGD Accuracy of ClusTR+QTRADES on more $\epsilon$ Values where we show that the robustness of the resultant model is agnostic from the choice of $\epsilon$.}
\centering
\begin{tabular}{l | cc cc  cc}
\hline \bottomrule  
 & \multicolumn{2}{c}{CIFAR10} & \multicolumn{2}{c}{SVHN} & \multicolumn{2}{c}{CIFAR100} \\
 $\epsilon$    & PGD$^{20}$ & PGD$^{100}$      & PGD$^{20}$ & PGD$^{100}$ & PGD$^{20}$ & PGD$^{100}$ \\ \midrule
\text{$\nicefrac{2}{255}$}              & 81.99 & 81.54           & 87.48           & 87.47 & 60.15 & 59.77 \\\hline
\text{$\nicefrac{16}{255}$}             & 57.67& 57.05           & 80.04 & 80.00& 33.32 & 33.25 \\\hline
\text{$\nicefrac{25.5}{255}$}    &             35.88 &  34.98               & 71.56 & 71.45& 17.76 & 17.65 \\
\hline \bottomrule      
\end{tabular}\label{tb:epsilons}
\end{table*}

Table \ref{tb:epsilons} reports the adversarial accuracies ClusTR + QTRADES under PGD attacks with $\epsilon \in \{\nicefrac{2}{255},\nicefrac{16}{255}, 0.1\}$ since we reported the results and comparisons for $\epsilon=\nicefrac{8}{255}$ in the main patper. Note that the robustness of our model is not limited to a specific value of $\epsilon$.

\subsection{Ablation on $D$}

\begin{figure}
    \centering
\includegraphics[width = \textwidth]{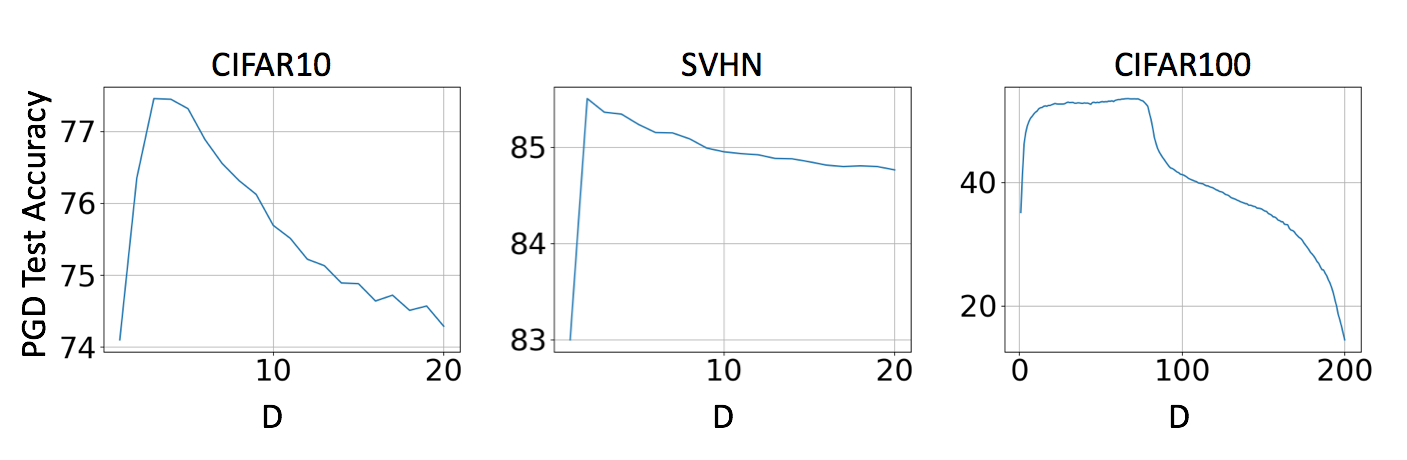}
    \caption{\textbf{Effect of $D$ on $\nicefrac{8}{255}-$PGD$^{20}$ Test Accuracy.} Note that with $D = 1$, \ie the assumption in our theoretical analysis, our methods outperforms the state-of-the-art. Moreover, it can be seen that considering only about $20\%$ of the total number of clusters yields the best performance.}
    \label{fig:ablation_on_D}
\end{figure}

ClusTR predicts the class of an input as a soft nearest cluster through Equation \eqref{eq:inference_ours}. The probabilities can also be computed by only considering the $D$ nearest clusters, as reported in the Experiments Section. Next, we report the effect of varying $D$ in terms of the natural and adversarial accuracies. 

Figure \ref{fig:ablation_on_D} depicts the behavior of clean and adversarial accuracies with varying $D$ on CIFAR10. We observe that the effect of varying $D$ on both CIFAR10 and SVHN is negligible $(\sim 3\%)$. The best PGD accuracy for both CIFAR10 and SVHN under the strong $\nicefrac{8}{255}-$PGD$^{20}$ attack was $77.04\%$ and $85.33\%$, respectively (corresponding to $D=4$). On the other hand, this effect seems to be stronger on CIFAR100. It is worthwhile to mention that more than 50\% of the choices of $D$ yields better robustness than the state of the art. Moreover, with $D=1$ which is exact setup of our theoretical result in Proposition \ref{theo:delta_bound}, ClusTR+QTRADES surpasses the state of the art on all of the datasets by a significant margin. Finally, the best $\nicefrac{8}{255}-$PGD$^{20}$ accuracy on CIFAR100 is 53.25\% with $D=60$.

\section{Proof of Proposition 1}
\emph{Proof.} It suffices that $\|f_\theta(x+\delta) - \mu_1\|^2 < \|f_\theta(x+\delta) - \mu_2\|^2$ for $x+\delta$ to be classified as $\mathcal{C}_1$. Therefore
\begin{equation}
\label{eq:interim res}
\begin{aligned}
  &  \|f_\theta(x+\delta) - \mu_2\|^2 - \|f_\theta(x+\delta) - \mu_1\|^2  \\
    &\qquad= \|f_\theta(x+\delta)-f_\theta(x)+f_\theta(x) - \mu_2\|^2 \\&\qquad- \|f_\theta(x+\delta)-f_\theta(x)+f_\theta(x)-\mu_1\|^2  \\
    & \qquad = \|f_\theta(x) - \mu_2\|^2 - \|f_\theta(x) - \mu_1\|^2 \\&\qquad+ 2\langle f_\theta(x+\delta) - f_\theta(x), \mu_1 - \mu_2 \rangle   \\ 
    & \ge \|f_\theta(x) - \mu_2\|^2 - \|f_\theta(x) - \mu_1\|^2 -2\mathcal{L} \|\delta\| \| \mu_2 - \mu_1\|. 
\end{aligned}
\end{equation}

The inequality follows by Cauchy-Schwarz and the Lipschitz property of $f_\theta$, \ie 
\begin{equation}
\begin{aligned}
    -\mathcal{L} \|\delta\| \| \mu_2 - \mu_1\|  &\leq |\langle f_\theta(x+\delta) - f_\theta(x), \mu_1 - \mu_2 \rangle| \\&\leq \mathcal{L} \|\delta\| \| \mu_2 - \mu_1\|.  \notag\\
\end{aligned}
\end{equation}
Thus, by rearranging the inequality in \ref{eq:interim res}, the bound on $\|\delta\|$ stated in Theorem~\ref{thm1} guarantees $\|f_\theta(x+\delta) - \mu_2\|^2 - \|f_\theta(x+\delta) - \mu_1\|^2 > 0$, completing the proof.

It is to be observed that the robustness radius is agnostic to the choice of $\mu_1$ and $\mu_2$. That is to say, the robustness radius in Theorem \ref{thm1} is not concerned with the accuracy of the classifier, but only with changes in the prediction under input perturbations. Therefore, the cluster centers $\mu_1$ and $\mu_2$ can be learnt jointly with the classifier's parameters $\theta$, such that the feature representations of inputs belonging to class $\mathcal{C}_1$ are close to some learnt $\mu_1$, while being far from the cluster center $\mu_2$ representing the other class. Note that if the clustering is performed, for example, with \emph{K-means}, then the cluster centers are the average features belonging to that class, \ie $\mu_i = \nicefrac{1}{|\mathcal{C}_i|}\sum_{x_j \in \mathcal{C}_i} f_\theta(x_j)$.

\end{document}